\documentclass[letterpaper]{article} 
\usepackage[submission]{aaai2026}  
\usepackage{times}  
\usepackage{helvet}  
\usepackage{courier}  
\usepackage[hyphens]{url}  
\usepackage{graphicx} 
\usepackage{natbib}  
\usepackage{caption} 
\usepackage[switch]{lineno}
\usepackage{algorithm}
\usepackage{algorithmic}
\usepackage{cite}
\usepackage{multirow}
\usepackage{makecell}
\usepackage{booktabs}
\usepackage[caption=false,font=footnotesize,labelfont=rm,textfont=rm]{subfig}
\usepackage{textcomp}
\usepackage{color}
\usepackage[switch]{lineno}

\usepackage{amssymb}
\usepackage{amsmath}
\usepackage{newfloat}
\usepackage{listings}
\DeclareCaptionStyle{ruled}{labelfont=normalfont,labelsep=colon,strut=off} 
\floatstyle{ruled}
\newfloat{listing}{tb}{lst}{}
\floatname{listing}{Listing}
\title{Adaptive Prototype Knowledge Transfer for Federated Learning with Mixed Modalities and Heterogeneous Tasks}
\author{
    Keke Gai\textsuperscript{\rm 1},
    Mohan Wang\textsuperscript{\rm 1},
    Jing Yu\textsuperscript{\rm 2},
    Dongjue Wang\textsuperscript{\rm 1},
    Qi Wu\textsuperscript{\rm 3}
}
\affiliations{
$^1$School of Cyberspace Science and Technology, Beijing Institute of Technology\\
$^2$School of Information Engineering, Minzu University of China\\
$^3$Australian Institute of Machine Learning, The University of Adelaide\\
 \{gaikeke,3120231264,3220231818\}@bit.edu.cn, jing.yu@muc.edu.cn, 
 qi.wu01@adelaide.edu.au
}

\usepackage{bibentry}

\begin{document}

\maketitle

\begin{abstract}

Multimodal Federated Learning (MFL) with mixed modalities enables unimodal and multimodal clients to collaboratively train models while ensuring clients' privacy.
As a representative sample of local data, prototypes offer an approach with low resource consumption and no reliance on prior knowledge for MFL with mixed modalities.
However, existing prototype-based MFL methods assume unified labels across clients and identical tasks per client, which is impractical in MFL with mixed modalities.
In this work, we propose an \underline{A}daptive \underline{pro}totype-based \underline{M}ultimodal \underline{F}ederated \underline{L}earning (AproMFL) framework for mixed modalities to address the aforementioned issues.
Our AproMFL transfers knowledge through adaptively-constructed prototypes without unified labels.
Clients adaptively select prototype construction methods in line with labels; server converts client prototypes into unified multimodal prototypes and cluster them to form global prototypes. 
To address model aggregation issues in task heterogeneity, we develop a client relationship graph-based scheme to dynamically adjust aggregation weights.
Furthermore, we propose a global prototype knowledge transfer loss and a global model knowledge transfer loss to enable the transfer of global knowledge to local knowledge.
Experimental results show that 
AproMFL outperforms four baselines on three highly heterogeneous datasets ($\alpha=0.1$) and two heterogeneous tasks, with the optimal results in accuracy and recall being $0.42\%\sim6.09\%$ and $1.6\%\sim3.89\%$ higher than those of FedIoT (FedAvg-based MFL), respectively.

\end{abstract}


%
\section{Introduction}
\label{sec:intro}
\begin{figure}[t]
\centering
\includegraphics[width=\columnwidth]{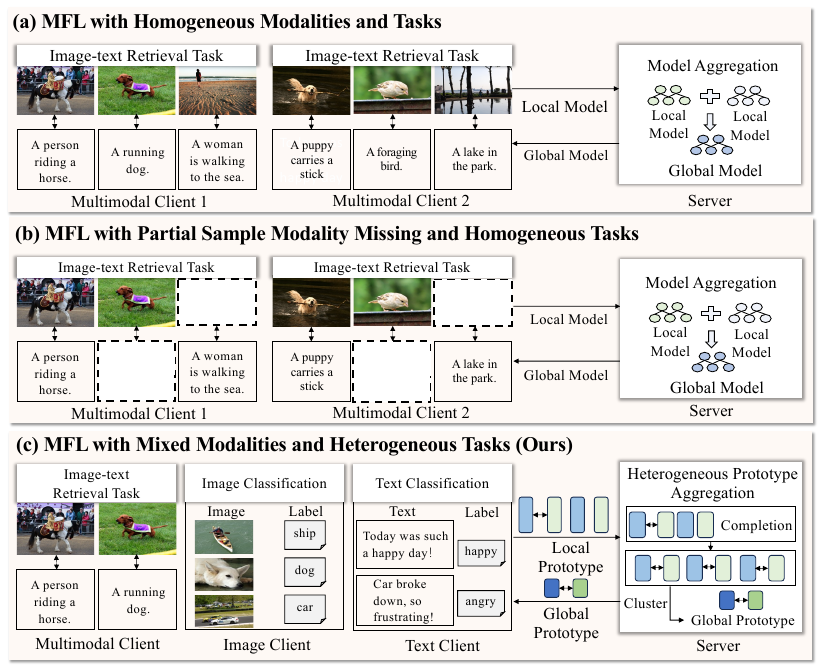} 
\vspace{-1em}
\caption{(a) The framework of MFL with homogeneous modalities and tasks. (b) The framework of MFL with partial sample modality missing and homogeneous tasks. (c) Adaptive prototype-based MFL framework for mixed modalities and heterogeneous tasks.}
\label{fig:moti}
\vspace{-2em}
\end{figure}

Multimodal Federated Learning (MFL)~\cite{feng2023fedmultimodal,chen2024feddat,li2024federated} has attracted increasing attention in recent years due to its technical merits in using multimodal data to collaboratively train models, which facilitates the extension of Federated Learning (FL) applications~\cite{huang2023rethinking}. 
Due to the advancement of hardware and network-related technologies, participants can collect data in multiple modalities, so that the traditional unimodal FL architecture no longer meets demands of collaborative model training for multimdoal clients ~\cite{wang2025pravfed}. 
Thus, attempts of MFL essentially aims at addressing the limitations caused by the assumption that each client has unimodal data and modalities across all clients are identical.

However, MFL still faces challenges deriving from mixed modallities.
Currently, varied sensing devices may cause modality heterogeneity, even though most previous studies~\cite{zong2021fedcmr,yan2024balancing,qi2024adaptive,li2023prototype} rely on a common assumption that all clients are modality-homogeneous (refer to Figure \ref{fig:moti}(a)).
Two common scenarios of heterogeneous modalities are partial sample modality missing~\cite{bao2023multimodal,xiong2023client} and mixed modalities~\cite{peng2024fedmm,sun2024towards}. 
Figure \ref{fig:moti}(b) exhibits a typical situation of partial sample modality missing, in which each client possesses a certain amount of multimodal data, guiding the alignment of locally incomplete modality samples. Most existing studies have focused on this scenario.
Figure \ref{fig:moti}(c) exhibits a typical case of mixed modality, in which multimodal data exist only in multimodal clients, while unimodal clients lack access to such data.
Effectively leveraging the knowledge from unimodal clients can further enrich the training data in MFL and enhance the model’s generalization ability.
However, the effective utilization of multimodal knowledge and the realization of modality knowledge transfer in mixed-modal MFL settings remain insufficiently explored.

Existing MFL methods with mixed modality can be divided into three types, including the public dataset-based methods ~\cite{yu2023multimodal,poudel2024car}, block-based methods ~\cite{chen2022fedmsplit}, prototype-based methods ~\cite{le2024cross} . 
The drawback of public dataset-based MFL is that the performance is highly dependent on the quality of public dataset.
Moreover, the public dataset must align with the semantic space of the clients' private data, yet privacy constraints make it difficult to access their semantic representations.
Block-based MFL tackles mixed modalities by dividing each model into modules to enable knowledge sharing through module aggregation. 
The challenge is that it involves all model components and causes a higher-level computational and communication overhead.
Prototypes, referring to the centroids of data from the same class, represent clients' local data characteristics. Compared with the previous two solutions, prototype-based MFL eliminates the need for public data and offers lower computational and storage costs.
However, these methods depend on an unpractical assumption that all clients' labels are unified and that each client performs the same task. In MFL with mixed modalities, the training tasks of multi-modal clients and those of unimodal clients are generally inconsistent, leading to difficult alignment of different samples and model drift~\cite{yu2023multimodal}.
In MFL with mixed modalities and heterogeneous tasks, constructing, aggregating, and utilizing prototypes to enable cross-client knowledge transfer without relying on consistent labels remains a pressing challenge.

To address above challenges, we propose an \underline{A}daptive \underline{pro}totype-based \underline{MFL} (AproMFL) framework for addressing issues of mixed modalities.
In AproMFL, neither the construction nor the aggregation of prototypes requires unified labels.
Clients adaptively select the manner of constructing local prototypes that represent local modal information, depending on the availability of local labels.
The server collects and aggregates clients' local prototypes. To aggregate multimodal and unimodal prototypes, it first performs semantic completion on unimodal ones to generate multimodal prototypes, then derives global prototypes via multimodal clustering. This process preserves cross-modal correspondences.
Moreover, to mitigate degradation caused from model averaging in task-heterogeneous, we develop a client relationship graph-based adaptive scheme for model aggregations. 
To reduce errors between local representations and global knowledge, our framework enables clients to use global multimodal prototype transfer loss and global model knowledge transfer loss for training local mapping modules, thereby strengthening local model generalization.

The main contributions are summarized as follows: 
(1) We propose a novel adaptive prototype-based MFL framework, AproMFL, which 
allows clients with heterogeneous modalities and tasks to participate in the FL training process. 
In AproMFL, neither the construction nor the aggregation of prototypes requires unified labels.
To the best of our knowledge, this is the first work to achieve MFL with mixed modalities and heterogeneous tasks through prototypes.
(2) We propose a cross-modal prototype aggregation scheme for matching demands of complex heterogeneous MFL, which allows the server to aggregate prototypes generated by different modalities and tasks.
(3) 
Our method outperforms four baselines on three highly heterogeneous datasets ($\alpha=0.1$) and two heterogeneous tasks, with the optimal results in accuracy and recall being $0.42\%\sim6.09\%$ and $1.6\%\sim3.89\%$ higher than those of FedIoT (FedAvg-based MFL), respectively.

\section{Related Work}
\label{sec:formatting}
\begin{figure*}[t]
\centering
\includegraphics[width=2\columnwidth]{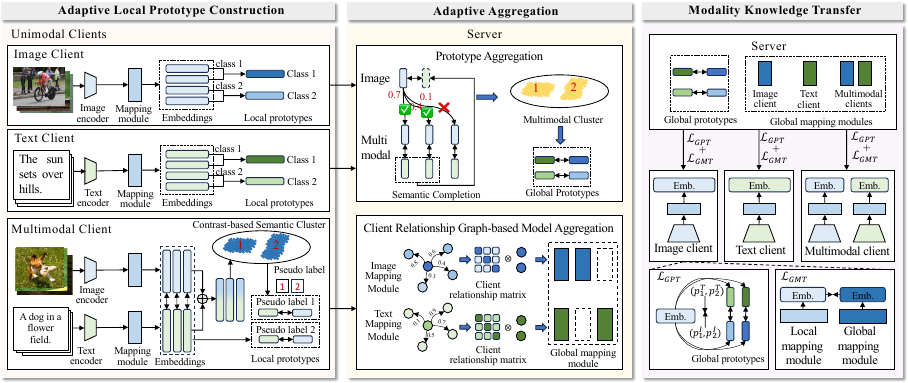} 
\caption{The framework of AproMFL. AproMFL consists of three modules: Adaptive local prototype construction builds local prototypes for different tasks; adaptive aggregation integrates multimodal local prototypes and models; modal knowledge transfer uses aggregated global knowledge to guide local model training.}
\label{fig:method}
\vspace{-1.5em}
\end{figure*}
\noindent\textbf{Data-Heterogeneous Federated Learning.}
FL is a distributed machine learning framework where clients train locally on private data and the server aggregates models to update a global one.~\cite{wang2024aggregation, yazdinejad2024robust}.
FedAvg~\cite{mcmahan2017communication} is one of the most representative algorithms. 
Some prior work has tried to address low performance due to data heterogeneity, where client data is non-independent and identically distributed (Non-IID)~\cite{li2023exploring}.
FedProx~\cite{li2020federated} and MOON~\cite{li2021model} introduce learning objectives to adjust the direction of local model training, while methods like FedAvgM~\cite{hsu2019measuring}, FedNova~\cite{wang2020tackling}, and FedMA~\cite{wang2020federated} mitigate heterogeneity's impact on model performance through aggregation.
Some methods~\cite{dai2023tackling,zhang2024fedtgp,zhou2025fedsa} alleviate data heterogeneity by extracting local prototypes and leveraging aggregated global prototypes to guide local training.
However, existing methods generally assume that clients are unimodal.
In MFL, clients may exhibit heterogeneity in modality, task, and statistics. 
Existing FL model aggregation and prototype construction methods for data heterogeneity typically consider only a single modality, overlooking cross-modal correlations, thereby limiting their applicability to MFL with mixed modalities.

\noindent\textbf{Multimodal Federated Learning.} MFL extends unimodal FL by enabling multimodal clients to participate in training~\cite{zong2021fedcmr,li2023prototype}.
Modality heterogeneity is a critical challenge in MFL, where clients differ in modality types. 
Current modality heterogeneity mainly falls into two types: MFL with partial modality missing in some samples~\cite{xiong2023client,bao2023multimodal} and MFL with mixed modalities~\cite{zhao2022multimodal,peng2024fedmm,le2024cross}.
Existing research primarily focuses on cases with partial modality missing in some samples, while the study of mixed modalities has not been fully explored. 
To address the challenge of aligning modality knowledge in mixed modalities, Yu et al.~\cite{yu2023multimodal} proposed an MFL framework that distills knowledge from clients with different modality types into a unified global model via knowledge exchange through a public dataset. Similarly, Huy et al.~\cite{le2024cross} developed a multimodal joint cross-prototype learning method that enables classification training with missing client modalities.
Sun et al.~\cite{sun2024towards} proposed a novel architecture named FedCola for Transformer models, which addresses both cross-modal and intra-modal discrepancies among clients.
However, these approaches rely on strong assumptions, such as the availability of a public dataset or sufficient labeled data for each client.

\section{Methodology}\label{method}

\subsection{Problem Formulation}

Assume that there are $M_M$ multimodal clients, $M_I$ image clients, $M_T$ text clients, and one server $S$.
Without loss of generality, we assume that multimodal clients have no labels, while unimodal clients possess labels. 
Each multi-modal client $c_i^M$($i \in [M_M]$) holds $N_i^M$ image-text pairs $\{(x_j^I, x_j^T)\}_{j=1}^{N_i^M}$, denoted by $D_i^M$. 
An image client \(c_i^I\)($i\in[M_I]$) possesses \(N_i^I\) images \(\{(x_j^I, y_j^I)\}_{j=1}^{N_i^I}\), denoted by \(D_i^I\). 
A text client \(c_i^T\)($i\in[M_T]$) holds \(N_i^T\) texts \(\{(x_j^T, y_j^T)\}_{j=1}^{N_i^T}\). 
\(y_j^I\) and \(y_j^T\) represent labels for images and texts, respectively.
We divide the model of each client $\Omega_i$ into an encoder $E_{i}^*, * \in \{I,T\}$ and a mapping module $f_i^*, * \in \{I,T\}$. 
The mapping module \(f_i^*\) with parameters \(\theta_i^*\) maps embeddings extracted by the encoders into a unified space. 
Clients participating in the classification task possess a classification module \(g_i^*\) with parameters \(w_i^*\) to obtain the final prediction output.
The objective function of AproMFL is defined by Equation (\ref{eq:obj}).
\begin{equation}\label{eq:obj}
    \min_{\{\Omega_i\}_{i=1}^M}\{R(\{\Omega_i\}_{i=1}^M)=\frac{1}{M}\sum_{i=1}^M\mathcal L_i(D_i^*,\Omega_i)\},
\end{equation}
where $\mathcal L_i$ denotes the loss for $c_i$, $M=M_M+M_I+M_T$.

AproMFL (refer to Figure~\ref{fig:method}) mainly consists of three components, including adaptive local prototype construction, server-side adaptive aggregation, and modality knowledge transfer.
By implementing our framework, clients achieve knowledge transfer across different modalities through prototypes without unified label information.

\subsection{Adaptive Local Prototype Construction}\label{ada-proto}
In MFL with mixed modality, unimodal and multi-modal clients handle distinct tasks. For instance, unimodal clients focus on classification tasks, while multi-modal clients concentrate on retrieval tasks. Given the heterogeneous label spaces across different tasks, local prototype construction cannot rely entirely on labels. However, complete abandonment of labels would also lead to information loss. To address this issue, we propose an adaptive local prototype construction method, which consists of two components: Label-guided Local Prototype Construction and Clustering-based Local Prototype Construction. Clients select specific construction methods according to whether they have labels.

\noindent\textbf{Label-guided Local Prototype Construction. }\label{sec:slocal}
We take an image client $c_i^I$ as an example to explain the label-guided construction, as the training process for text clients is similar.
Local image data are mapped into a unified space to obtain image embedding $e_i^I$, and the process is conceptualized by $ e_j^I = f_i^I\bigl(E_i^I(x_j^I),\theta_i^I\bigr)$.
Let the number of classes for $c_i^I$ is $\mathcal C_i^I$. 
Equation (\ref{eq:lproto}) defines the prototype of the $k$-th class.
\begin{equation}\label{eq:lproto}
    p_I^k = \frac{1}{|D_{ik}^I|}\sum_{j\in D_{ik}^I } e_j^I,
\end{equation}
where $D_{ik}^I$ denotes the subset of samples corresponding to the $k$-th class in the dataset $D_{i}^I$. 
To obtain prototypes with better representations of local data, we use the task loss $\mathcal L_{task}$, global prototype knowledge transfer loss $\mathcal L_{GPT}$ and global model knowledge transfer loss $\mathcal L_{GMT}$ to guide learning of the client model.
We utilize the task loss to guide the model in learning task-related features, 
i.e., a cross-entropy loss for classification tasks and a contrastive loss for multimodal retrieval tasks. 
$\mathcal L_{GPT}$ and $\mathcal L_{GMT}$ are used to align local and global knowledge (see the modality knowledge transfer section).
After multiple training rounds, the client computes local prototypes by Equation (\ref{eq:lproto}) and sends the local prototypes and 
the mapping module 
to the server.

\noindent\textbf{Clustering-based Local Prototype Construction. }\label{sec:ulpc}
Distinct from label-guided construction, multi-modal clients acquire local multi-modal prototype pairs through clustering. 
In the process of prototype construction, we still retain the semantic information of multi-modal prototype pairs. Specifically,
multimodal client $c_i^M$ obtains image-text embeddings pairs $(e_j^I,e_j^T)$ when inputting samples of multiple modalities into the mapping module and fusing modalities' embeddings, denoted by $e_j^M$, where $e_j^M={(e_j^I+e_j^T)}/{2}$. 
Unlike other existing methods that typically use cluster centroids as local prototypes, we use the $K$-means to cluster the fused embeddings and obtain pseudo-labels.
After obtaining the pseudo-labels, we construct the set $S_k^P$ for $k \in [K]$, which contains all image-text embedding pairs with the same pseudo-label $k$; we compute the mean of image embeddings and the mean of text embeddings within each set, making the mean pair set ${(p_i^I, p_i^T)}_{i=1}^{K}$ the local prototype set for multimodal clients.

To enhance clustering effectiveness, 
we involve a group of losses in the clustering model training, such as the task, intra-modal contrastive, and inter-modal contrastive loss.
Specifically, the task loss is similar to per discussed. 
The intra-modal contrastive loss is used to make those embeddings with the same pseudo-label in the same modality (positive samples) becoming closer.
The intra-modal contrastive loss for the $i$-th sample is defined as follows,
\begin{equation}\label{lintra}
    \mathcal L^{intra}_{i,*}=-\frac{1}{|S_k^P|}\sum_{j\in S_k^P}\log \frac{\exp (S(e_i^*,e_j^*)/\tau)}{\sum_{t=1}^{N^M}\exp (S(e_i^*,e_t^*)/\tau) },
\end{equation}
where $*\in\{I,T\}$. 
A better clustering can be achieved as minimizing the intra-modal contrastive loss sharpens the boundary between positive and negative samples.
The inter-modal contrastive loss is defined as follows,
\begin{equation}\label{linter}
    \mathcal L^{inter}_i=-\frac{1}{|S_k^P|}\sum_{j\in S_k^P}\log \frac{\exp (S(e_i^I,e_j^T)/\tau)}{\sum_{t=1}^{N^M}\exp (S(e_i^I,e_t^T)/\tau) }
\end{equation}
The inter-modal contrastive loss aligns image-text embeddings with the same pseudo-label. 
The overall loss function $\mathcal L_M$ during the clustering process is defined as follows,
\begin{equation}\label{utloss}
    \mathcal L_M = \mathcal L_{task}+\sum_{i=1}^{N^M}(\sum_{*\in\{I,T\}} L^{intra}_{i,*}+\mathcal L^{inter}_i).
\end{equation}
After training the clustering model, multi-modal clients only send local prototype pairs to the server for aggregation, while the clustering model remains local.

To learn embeddings with global knowledge across different modalities, we train the local task model by using $\mathcal L_{task}$, $\mathcal L_{GPT}$, $\mathcal L_{GMT}$ and local mapping module regularization loss $\mathcal L_{LMR}$. 
The objective of the $\mathcal L_{LMR}$ is to minimize the difference between the mapping module of the task model and that of the clustering model, defined as follows,
\begin{equation}\label{eq:lu}
    \mathcal L_{LMR}=\lambda\sum_{*\in\{I,T\}}||\theta^*-\theta_p^*||_2^2,
\end{equation}
where $\theta_p^*$ represents the private mapping module obtained from clustering model training. 
$\lambda$ is a parameter used to balance the relationship between personalization and global knowledge. 
$\mathcal L_{LMR}$ not only accelerates the task model training but also facilitates knowledge distillation from the existing private mapping module into the task model.
The trained task model is sent to the server for aggregation.

\subsection{Server-side Adaptive Aggregation}\label{agg}
This component addresses limitations deriving from the implementation of averaging aggregation in heterogeneous modalities and tasks. 
Two key aggregations are involved.

\noindent\textbf{Adaptive Heterogeneous Prototype Aggregation.}
The prototypes received by the server include both unimodal and multimodal prototypes.
The aggregation includes semantic completion and multimodal clustering. Taking the image client $c_i^I$ as an example, for semantic completion, the server first computes the cosine similarity between image prototype $c_i^I$ and image prototypes of the multimodal clients. 
The top-$\mathcal O$ most multimodal image prototypes are selected, denoting corresponding text prototypes as $\tilde{P}_t = \{p_t^o\}_{o=1}^{\mathcal O}$. 
We denote $\mathcal O$ as the modality completion parameter.
Thus, we convert similarities into weight values, by applying a positive relationship between similarity and weight.
Then, new image-text prototype pairs are formed from the obtained prototypes paired with the image prototype of $c_i^I$, i.e., multiplying each element in $\tilde{P}_t$ by its corresponding weight.
The same operations also apply for text clients prototypes. 
Server aggregates local prototypes by a multimodal clustering once obtaining the prototypes. 
The process is similar to the construction of local prototypes by multimodal clients. 
Eventually, the server obtains $K$ prototype pairs.

\noindent\textbf{Client Relationship Graph-based Model Aggregation.}
To address issue of varied feature spaces of clients' local models, we propose client relationship graph-based model aggregation scheme that aims at mitigating performance degradation by only aggregating each client's mapping modules. 
Take the image client $c_i^I$ as an example.
The similarity between client's image mapping module and those of other clients is computed. Similarity between client $i$ and $j$ is defined as  $s(i,j)={\theta_i\cdot\theta_j}/{||\theta_i|| \cdot||\theta_j||}$, where $\theta$ is the parameter of mapping module. 
These similarities are normalized into weights, and eventually each image mapping module is weighted accordingly, making an aggregated image mapping module for the client. 
Text clients follow the same aggregation process, while multimodal clients separately aggregate image and text mapping modules using their corresponding similarities.

\begin{table*}[th!]
	\centering
	\begin{tabular*}{0.968\linewidth}{c|c|cc|cc|cccc}
		\bottomrule
		\multirow{2}*{Methods} & \multirow{2}*{$\alpha$}&\multicolumn{2}{c|}{CIFAR-10} & \multicolumn{2}{c|}{AG-NEWS} &\multicolumn{4}{c}{Flickr30k}\\
        \cline{3-10}
      & & Acc@1 &Acc@5&Acc@1 &Acc@2&R@1 (i2t)&R@5 (i2t) &R@1 (t2i)&R@5 (t2i)\\
		\hline 
       \multirow{2}*{Local}
        & 0.1 & 57.21\% & 65.03\%& 57.59\%& 59.43\%& 48.00\% & 77.44\% & 36.04\% & 70.19\% \\
        & 5.0 & 90.77\% & 93.79\%& 91.14\%& 97.73\%& 48.90\% & 77.79\% & 35.10\% & 69.37\% \\
        \hline
         \multirow{2}*{FedIoT}
        & 0.1 & 57.68\% & 64.97\%& 58.72\%& 62.23\%& 48.29\% & 77.15\% & 35.58\% & 69.56\%  \\
        & 5.0 & 89.75\% & 94.84\%& 90.26\%& \textbf{98.38\%}& 46.89\% & 77.09\% & 35.26\% & 69.66\% \\
        \hline
         \multirow{2}*{MM-FedProx}
        & 0.1 & 58.26\% & 64.61\%& 56.99\%& 62.34\%& 42.04\% & 71.34\% & 31.16\% & 65.18\% \\
        & 5.0 & 91.29\% & 94.74\%& 54.52\%& 73.65\%& 42.79\% & 73.74\% & 33.42\% & 67.30\% \\
        \hline
        \multirow{2}*{CreamFL}
        & 0.1 & 20.62\% & 49.24\%& 25.01\%& 50.93\%& 39.26\%&68.42\% & 29.24\% & 56.52\% \\
        & 5.0 & 25.01\%	 & 55.56\%	& 38.66\%	& 49.07\%& 42.66\% & 70.13\% & 32.10\% & 58.98\% \\
        \hline
        \multirow{2}*{AproMFL (Ours)}
        & 0.1 & \textbf{59.71\%} & \textbf{67.37\%}& \textbf{59.14\%}& \textbf{68.32\%}& \textbf{50.09\%} & \textbf{78.75\%} & \textbf{39.47\%} & \textbf{71.86\%} \\
        & 5.0 & \textbf{96.16\%} & \textbf{99.68\%}& \textbf{92.07\%}& 97.00\%& \textbf{50.90\%} & \textbf{78.65\%} & \textbf{39.45\%} & \textbf{71.86\%} \\
        
       \toprule
       
	\end{tabular*}
    \vspace{-0.5em}
    \caption{Comparison of the average accuracy or recall of models across different methods under varying degrees of data heterogeneity.}
 \label{baseline_sm}
 \vspace{-2em}
\end{table*}

\subsection{Modality Knowledge Transfer}\label{sec:modality_KT}

To align global modality knowledge with local modality knowledge, we utilize global prototype pairs and the global model to guide local model training.
We adopt a global model knowledge transfer loss $\mathcal L_{GPT}$ and a global prototype knowledge transfer loss $\mathcal L_{GKT}$ during the client's local training process.
For $\mathcal L_{GPT}$, we denote the global prototype as $\hat P_g = \{(p_i^I,p_i^T)\}_{i=1}^{K}$.
Taking the image client as an example, for the $j$-th image sample, the client calculates the assignment probability of the image embedding $e_j^I$ to the $i$-th global image prototype. 
The assignment probability is defined as follows,
\begin{equation}\label{eq:q}
    q_{j,i}^I = \frac{\exp(\frac{1}{\tau}S(e_j^I,p_i^I))}{\sum_{l=1}^{\mathcal K}\exp(\frac{1}{\tau}S(e_j^I,p_l^I))},
\end{equation}
where $S(;,;)$ denotes the cosine similarity. For $K$ global image prototypes, we ultimately obtain $K$ assignment probabilities, denoted as $Q_j^I=(q_{j,1}^I,...,q_{j,\mathcal K}^I)$.  Similarly, we obtain the assignment probabilities for the local image embedding $e_j^I$ to the $K$ global text prototypes, denoted as $Q_j^T=(q_{j,1}^T,...,q_{j,\mathcal K}^T)$.
Since the global image-text prototypes are paired, we assume the assignment probabilities of the local image embedding to the paired image and text prototypes should be aligned. The $\mathcal L_{GPT}$ is defined as follows,
\begin{equation}
\resizebox{0.95\hsize}{!}{$
  \begin{aligned}\label{eq:js}
    &  \mathcal L_{GPT} 
            =\frac{1}{2}D_{KL}(Q_j^I||(\frac{Q_j^I+Q_j^T}{2}))  + \frac{1}{2}D_{KL}(Q_j^T||(\frac{Q_j^I+Q_j^T}{2})),
  \end{aligned} 
  $}
\end{equation}
where 
$D_{KL}$ represents the Kullback-Leibler divergence.

To further reduce the deviation between local model and global model, we adopt a loss $\mathcal L_{GMT}$.
After receiving a global model, client uses it as a teacher model for knowledge distillation, such that the embeddings ($Emb_l^I$) output by the local model are made to align with those ($Emb_g^I$) output by the global model. 
To prevent a poorly performing global model from affecting local model training, we adopt a factor $\nu$ ($\nu=\mathcal L_{task}^l/\mathcal L_{task}^g$).
When a task loss of current local model ($\mathcal L_{task}^l$) is smaller than that of the global model ($\mathcal L_{task}^l$), the factor reduces the amount of knowledge transferred from the global model to the local model, and vice versa. 
The $\mathcal L_{GMT}$ is defined as follows,
\begin{equation}
  \begin{aligned}\label{eq:gmt}
    &  \mathcal L_{GMT} = \nu D_{KL}(Emb_l^I||Emb_g^I).
  \end{aligned} 
\end{equation}

Computations of $\mathcal{L}_{GPT}$ and $\mathcal{L}_{GMT}$ for image and multimodal clients are similar. 
The difference is that $Q_j^T$ represents the allocation probability of the j-th text embedding to each global text prototype for multimodal clients. 
Under the guidance of these two losses, global modality knowledge is effectively transferred to local clients.

\begin{algorithm}[t]
\small
\caption{\bf {Federated Training of AproMFL}}
\label{alg1}
\begin{algorithmic}
\REQUIRE {Client $c_i^M (i\in [M_M])$, $c_i^I(i\in [M_I])$, $c_i^T(i\in[M_T])$, Multimodal datasets $D^M$, Unimodal datasets $D^I,D^T$}
\end{algorithmic}
\begin{algorithmic}[1]
\STATE Each client initializes the local model $\Omega$. 
\FOR {each round $t=1,...,\mathbb T$}
\FOR {each unimodal client $c_i^*,*\in[I,T]$}
\FOR{each local round $r=1,...,\mathbb R$} 
\STATE Update the local model $\Omega_i$ 
\ENDFOR
\STATE Calculate the prototype according to Equation \ref{eq:lproto}.
\ENDFOR
\FOR {each multimodal client $c_i^M$}
\STATE Construct local prototype pairs via clustering. 
\FOR{each local round $r=1,...,\mathbb R$} 
\STATE Compute $\mathcal L_{task},\mathcal L_{GPT},\mathcal L_{GMT},\mathcal L_{GMR}$.
\STATE Update the local model $\Omega_i$ 
\ENDFOR
\STATE Send the local prototype and the mapping module $\theta_i^*,*\in{I,T}$ to the server.
\STATE //Server
\STATE Aggregate the clients' prototypes to derive the global prototype.
\STATE Aggregate local models according to the client relationship graph. 
\STATE Send the aggregated model and prototypes to the clients.
\ENDFOR
\ENDFOR
\end{algorithmic}
\end{algorithm}

\section{Experiments}
\label{sec:rationale}

\begin{figure*}[t]
  \centering

   \subfloat[Local (Image)]
   {\includegraphics[width=0.24\textwidth]{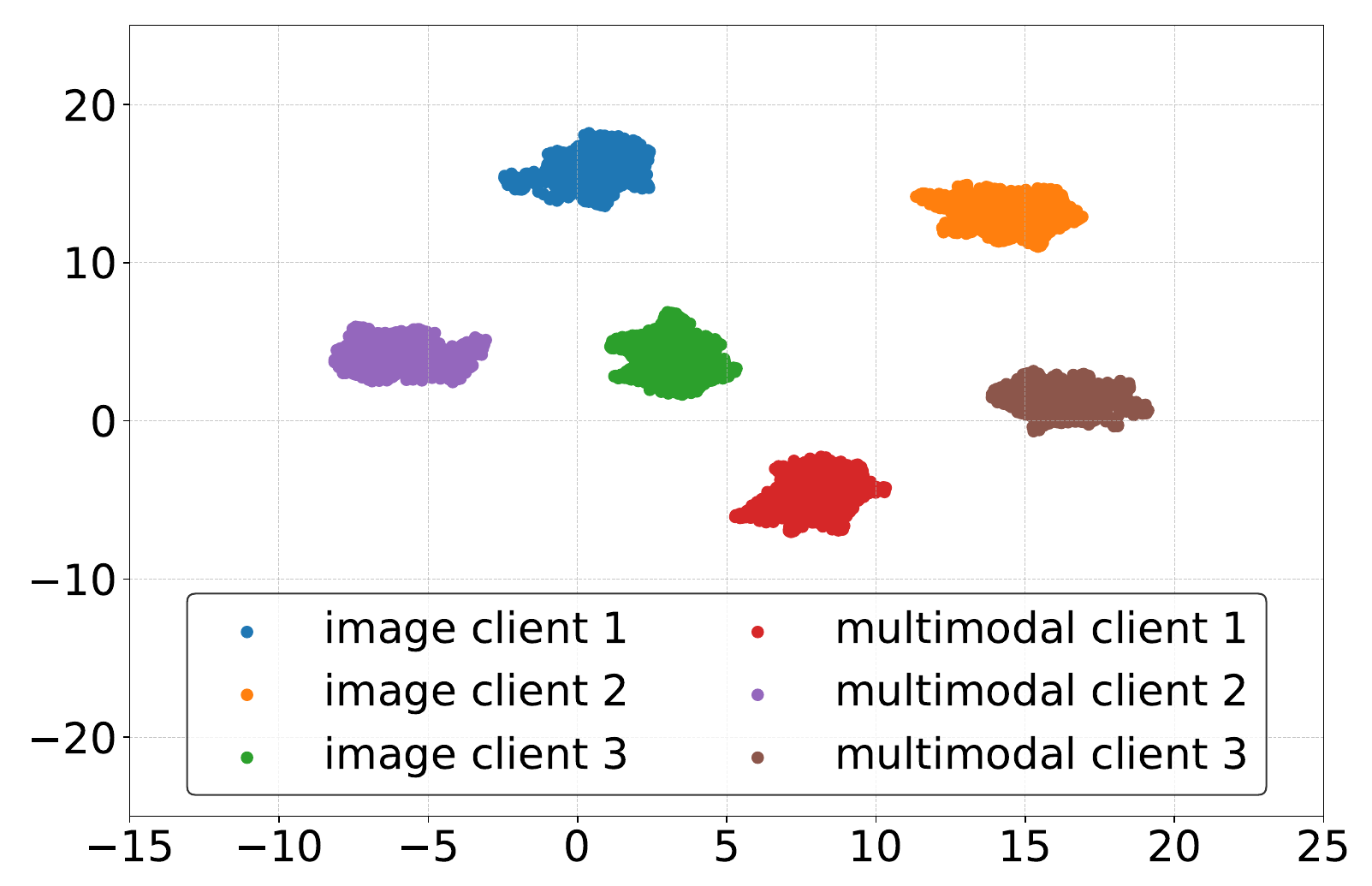}\label{fig-mmalign:suba}}
  \hfil
  \subfloat[AproMFL (Image)]
  {\includegraphics[width=0.24\textwidth]{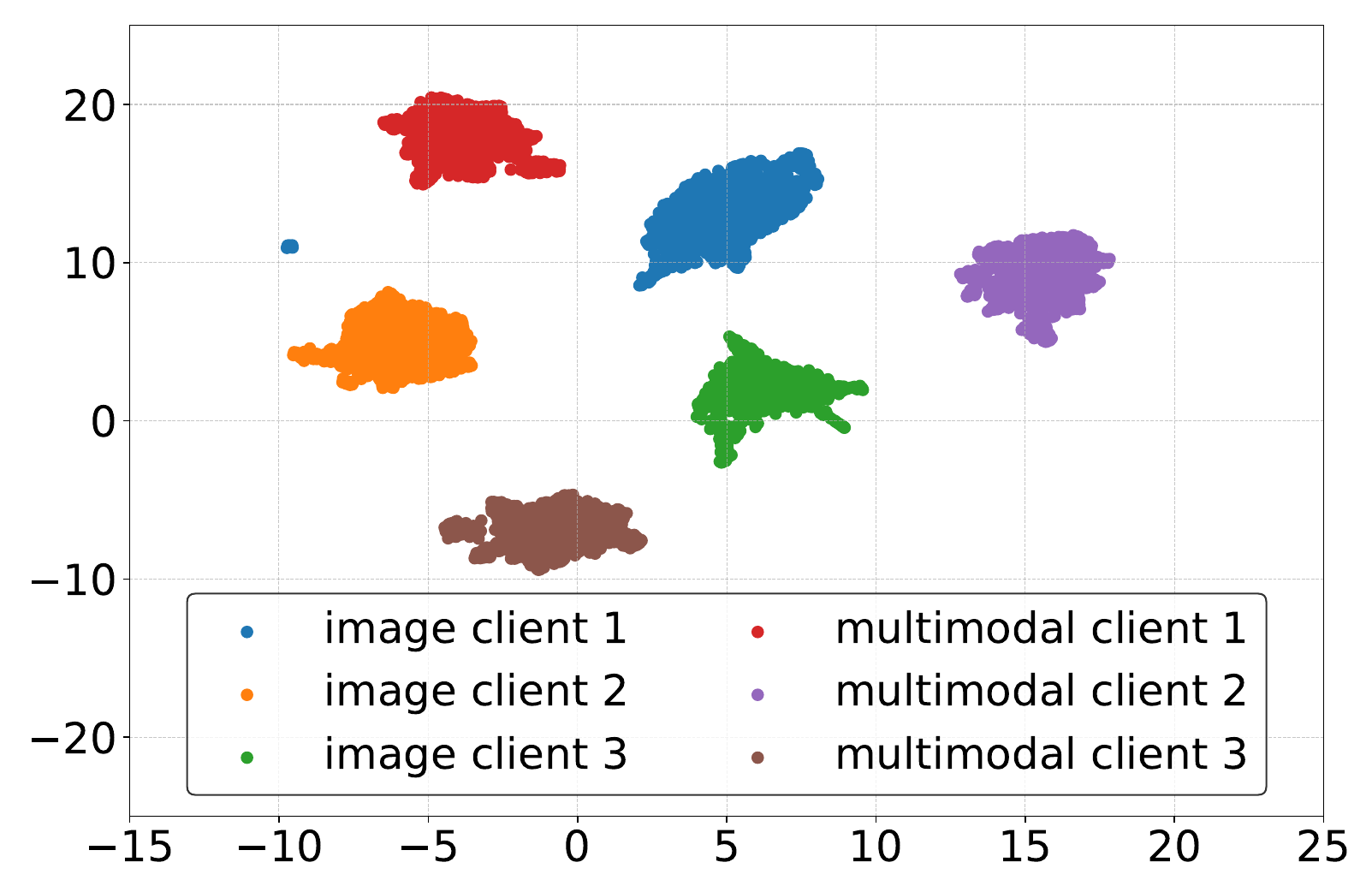}\label{fig-mmalign:subb}}
  \hfil
  \subfloat[Local (Text)]
  {\includegraphics[width=0.24\textwidth]{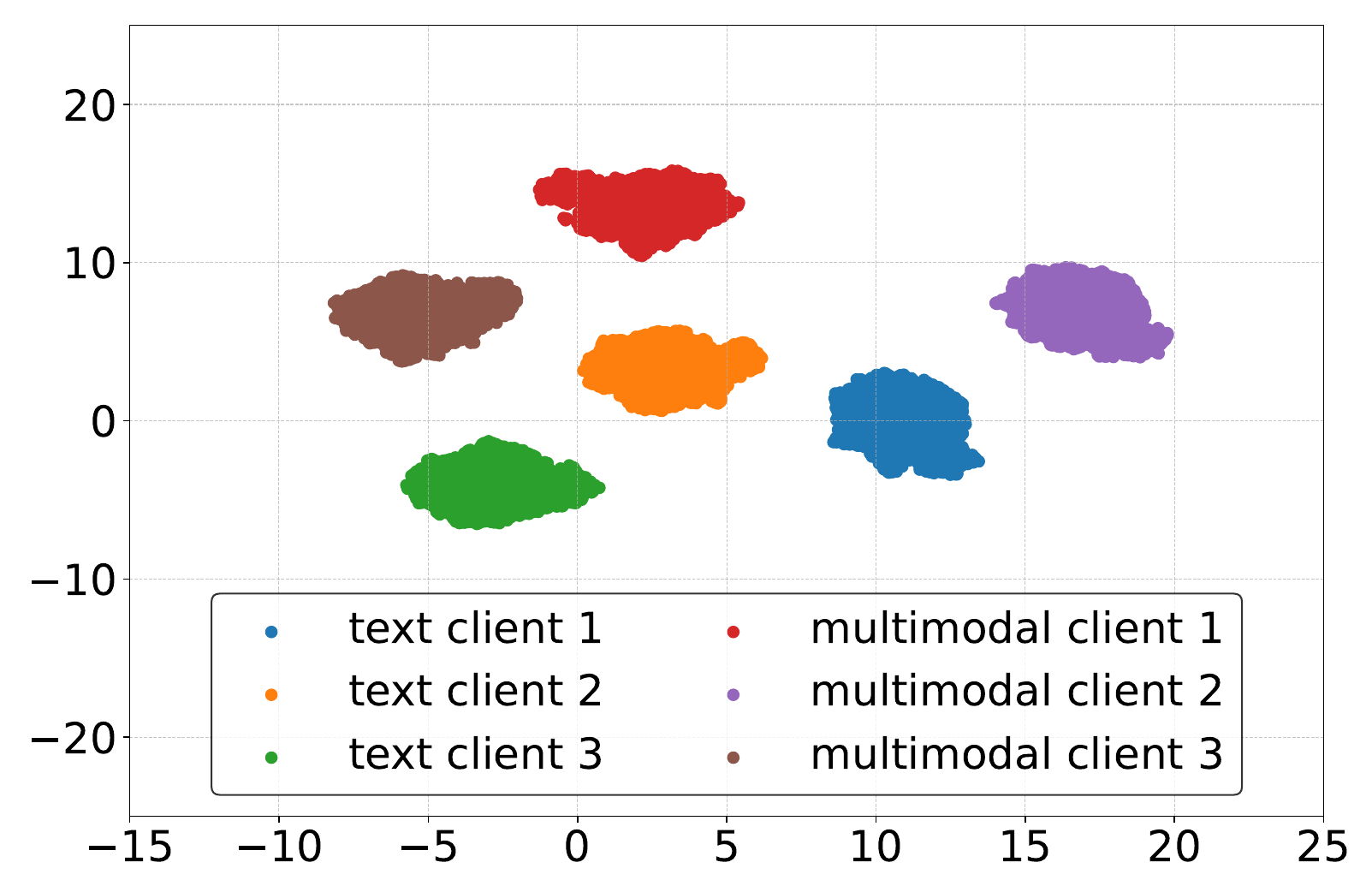}\label{fig-mmalign:subc}}
  \hfil
  \subfloat[AproMFL (Text)]
  {\includegraphics[width=0.24\textwidth]{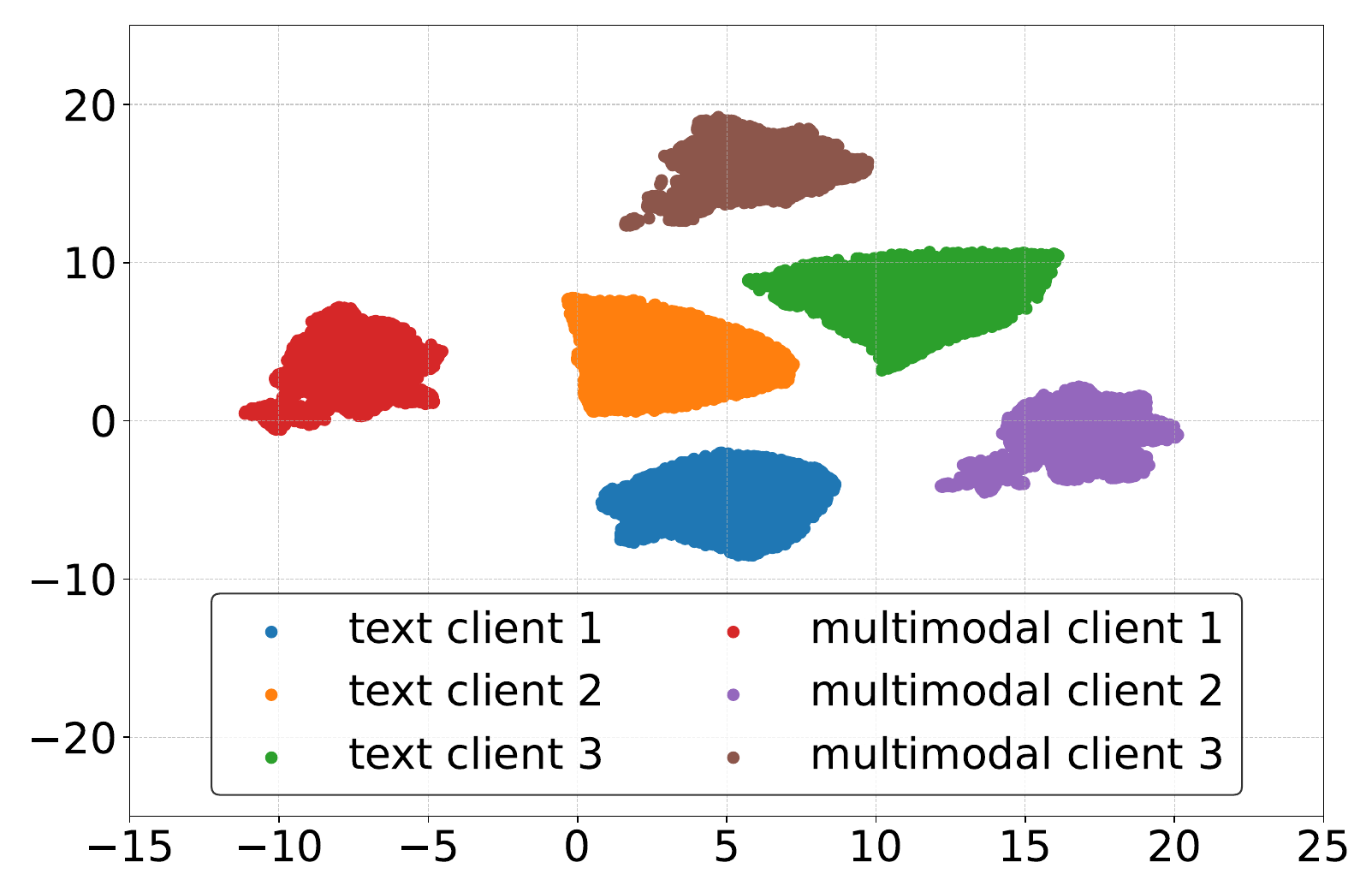}\label{fig-mmalign:subd}}
\vspace{-0.5em}
  \caption{Distribution of client representations under the Flickr30k dataset.}
  \label{fig-mmalign}
 \vspace{-2em}
\end{figure*}

\noindent\textbf{Datasets.}
We evaluate the performance of AproMFL in heterogeneous modality scenarios across three datasets, including CIFAR-10~\cite{krizhevsky2009learning}, AG-NEWS~\cite{zhang2015character}, and Flickr30k~\cite{young2014image}. 
We allocate the CIFAR-10 to the image clients, the AG-NEWS to the text clients, and the Flickr30k to the multimodal clients. 
To simulate Non-IID data, we use the Dirichlet distribution for data partitioning~\cite{hsu2019measuring}, i.e., a smaller value of $\alpha$ indicates a higher-level data heterogeneity.

\noindent\textbf{Implementation Details.}
We use the CLIP~\cite{radford2021learning} ViT-L/14 as encoders for different clients. 
The mapping module of each client is configured as a three-layer Fully Connected Network (FCN).
In multimodal retrieval tasks, we test both image-to-text (i2t) and text-to-image (t2i) retrieval, measuring the top-1 and top-5 recall, denoted by R@1 and R@5, respectively.
In classification tasks, we test both top-1 and top-5 accuracy for CIFAR-10, denoted by Acc@1 and Acc@5, respectively. We also test both top-1 and top-2 accuracy for AG-NEWS, denoted by Acc@1 and Acc@2, respectively.
All experiments are on an RTX 4090 GPU, with Python 3.9, PyTorch 2.2.2, and CUDA 11.8.

\noindent\textbf{Baselines.}
We compared AproMFL with existing MFL methods, covering followings. 
(1) Local, a method that considers only local training on clients without modality knowledge sharing.
(2) FedIoT~\cite{zhao2022multimodal}, an extension of FedAvg~\cite{mcmahan2017communication} adapted to the multimodal federated setting. 
(3) MM-FedProx, an adaptation of FedProx~\cite{li2020federated}, originally designed to handle data heterogeneity in unimodal scenarios, here extended to the multimodal federated setting.
(4) CreamFL~\cite{yu2023multimodal} is a knowledge distillation-based MFL framework that facilitates knowledge sharing among clients with different modalities by transmitting low-dimensional representations of a public dataset.


\subsection{Comparison with State-of-the-Art Methods}
Table \ref{baseline_sm} presents a comparison of the average accuracy and recall of client models across different methods under varying degrees of data heterogeneity.
The highest accuracy and recall have been bolded.
Except for Acc@2 ($\alpha=5.0$), AproMFL outperforms the baselines in terms of both accuracy and recall.
Specifically, in the classification task, AproMFL achieves an Acc@1 ($\alpha = 0.1$) that is 2.50\% and 1.55\% higher than local on the CIFAR-10 and AG-NEWS.
In retrieval tasks, AproMFL achieves a R@1 that is 2.09\% and 3.43\% higher than local on i2t and t2i.
This indicates that our scheme effectively combines data from different clients to produce better-performing models. 
Furthermore, AproMFL outperforms FedIoT, MM-FedProx, and CreamFL in both classification and retrieval tasks, which evidence that our scheme effecitvely leverages knowledge from clients with different modalities, achieving a great improvement on downstream tasks.

\begin{table}[t]
	\centering
	\begin{tabular}{c|c|ccc}
		\bottomrule
        Datasets & Metrics & w/o GP & w/o GM & AproMFL \\ \hline
	  \multirow{1}*{CIFAR-10}& Acc@1& 59.69\%&58.73\%	&59.71\% \\
        \hline
      \multirow{1}*{AG-NEWS}&Acc@1& 58.75\%&58.94\%	&59.14\% \\
      \hline
       \multirow{2}*{Flickr30k}&{R@1}$_s$&88.75\% & 88.55\%& 89.36\%\\
       &R@5$_s$&150.75\% & 150.59\%& 150.61\%\\
      
       \toprule
       
	\end{tabular}
    \vspace{-0.5em}
    \caption{The comparison of accuracy and recall between AproMFL w/o GP, AproMFL w/o GM, and AproMFL.}
 \label{ablation}
 \vspace{-2em}
\end{table}

\subsection{Ablation Study}
We investigate the impact of global prototypes and global models on model performance. 
AproMFL w/o GP refers to the AproMFL without global prototypes knowledge transfer loss, while AproMFL w/o GM refers to the AproMFL without global models knowledge transfer loss. 
R@1$_s$ denotes the sum of top-1 recall for the i2t and t2i tasks, while R@5$_s$ represents the sum of top-5 recall for these tasks.
Table \ref{ablation} presents the test accuracy and recall for the three methods.
We compare AproMFL w/o GP and AproMFL w/o GM with Local and FedIoT from Table \ref{baseline_sm}. 
It is observed that both AproMFL w/o GP and AproMFL w/o GM outperform Local in terms of accuracy and recall across all three datasets. This suggests that models guided by either global prototypes or global models perform better than those trained locally. Additionally, AproMFL w/o GP exhibits superior accuarcy and recall compared to FedIoT. This implies that global models derived through model-adaptive aggregation are more effective in minimizing discrepancies between clients and promoting knowledge sharing, compared to models obtained via FedAvg. 
AproMFL performs best in Acc@1 and R@1, highlighting that combining global prototypes and models to regularize local training enhances model performance.

\subsection{Performance of Modality Alignment}
We evaluate the alignment of modality knowledge in AproMFL using the Flickr30k datasets.
We set $\alpha=0.1$, with three clients for each type. The test set images and text data from Flickr30k are input into the image and text models trained via Local and AproMFL, respectively. The resulting embeddings are visualized after dimensionality reduction using UMAP~\cite{Healy2024}.
As shown in Figure \ref{fig-mmalign}, on the same data, Local-derived embeddings exhibit stronger spatial bias toward clients' local knowledge, while AproMFL-generated embeddings are more spatially compact. This demonstrates that AproMFL can incorporate global knowledge while retaining client-specific expertise, facilitating consistent representation learning across clients.



\begin{table}[t!]
	\centering
    	\begin{tabular}{c|c|c|cc}
		\bottomrule
		\multirow{2}*{K} & \multicolumn{1}{c|}{CIFAR-10} & \multicolumn{1}{c|}{AG-NEWS} &\multicolumn{2}{c}{Flickr30k}\\
        \cline{2-5}
      & Acc@1 &Acc@1 &R@1$_s$ &R@5$_s$\\
		\hline 
       10 & 59.71\% & 59.14\%& 89.56\% & 150.61\% \\
       20 & 59.36\% & 59.57\%& 89.02\% & 151.08\% \\
       40 & 60.36\% & 59.57\%& 87.07\% & 150.90\% \\
       60 & 60.81\% & 59.47\%& 86.95\% & 150.83\% \\
       80 & 60.08\% & 59.59\%& 86.12\% & 150.21\% \\
       \toprule       
	\end{tabular}
\vspace{-0.5em}
    \caption{Average accuracy and recall of the model under different numbers of prototypes.}
 \label{proto_num}
 \vspace{-1.5em}
\end{table}

\begin{table}[t!]
	\centering
    	\begin{tabular}{c|c|c|cc}
		\bottomrule
		\multirow{2}*{$\mathcal O$} & \multicolumn{1}{c|}{CIFAR-10} & \multicolumn{1}{c|}{AG-NEWS} &\multicolumn{2}{c}{Flickr30k}\\
        \cline{2-5}
      & Acc@1 &Acc@1 &R@1$_s$&R@5$_s$ \\
		\hline 
        2 &60.98\% & 59.16\%&  90.33\% & 151.16\% \\
       5 & 60.50\% & 58.49\%&  90.37\% & 150.99\% \\
       8 &60.93\% & 59.57\%& 90.22\% & 151.07\% \\
       10 & 59.71\% & 59.14\%& 89.56\% & 150.61\% \\
       \toprule       
	\end{tabular}
\vspace{-0.5em}
    \caption{Average accuracy or recall of the model under different modality completion parameter $\mathcal O$.}
 \label{completion_o}
 \vspace{-1.5em}
\end{table}
\subsection{Parameter Analysis}
\noindent\textbf{Performance under Different Numbers of Prototypes.}
Table \ref{proto_num} presents the model's average accuracy and recall for different numbers of global prototype pairs. 
As shown in Table \ref{proto_num}, with the increase in the number of global prototypes, our method's accuracy on CIFAR-10 and AG-NEWS, as well as recall on Flickr30k, fluctuates within a certain range.
For multimodal clients, as the number of global prototypes increases, the changes in R@1$_s$ and R@5$_s$ are 3.44\% and 0.87\%, respectively. For unimodal clients, 
the changes in Acc@1 for CIFAR-10 and AG-NEWS are 1.45\% and 0.45\%, respectively
This shows that our scheme is robust to different numbers of global prototypes.

\begin{figure}[t]
  \centering

   \subfloat[Classification]
  {\includegraphics[width=0.23\textwidth]{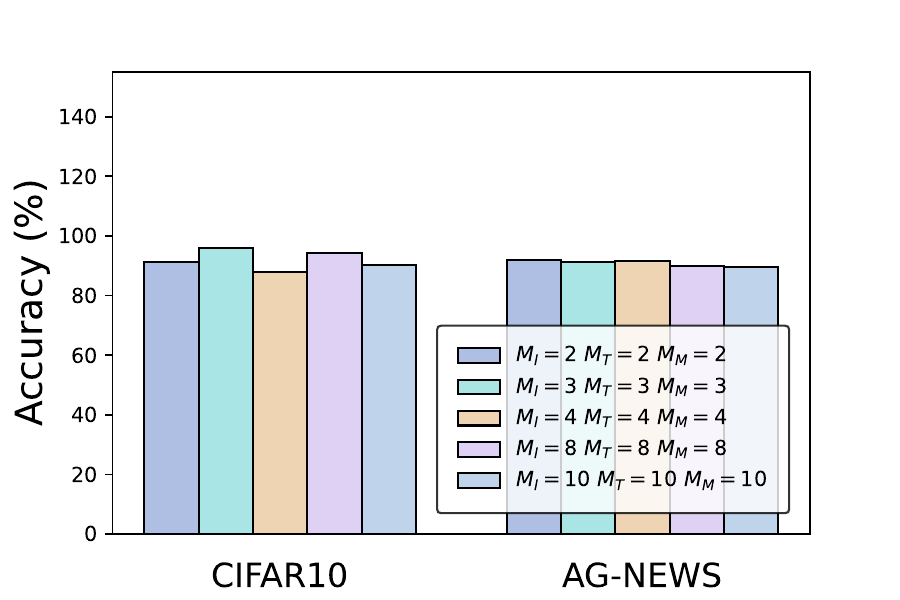}\label{fig3:subc}}
  \hfil
  \subfloat[Multi-modal Retrieval]
  {\includegraphics[width=0.23\textwidth]{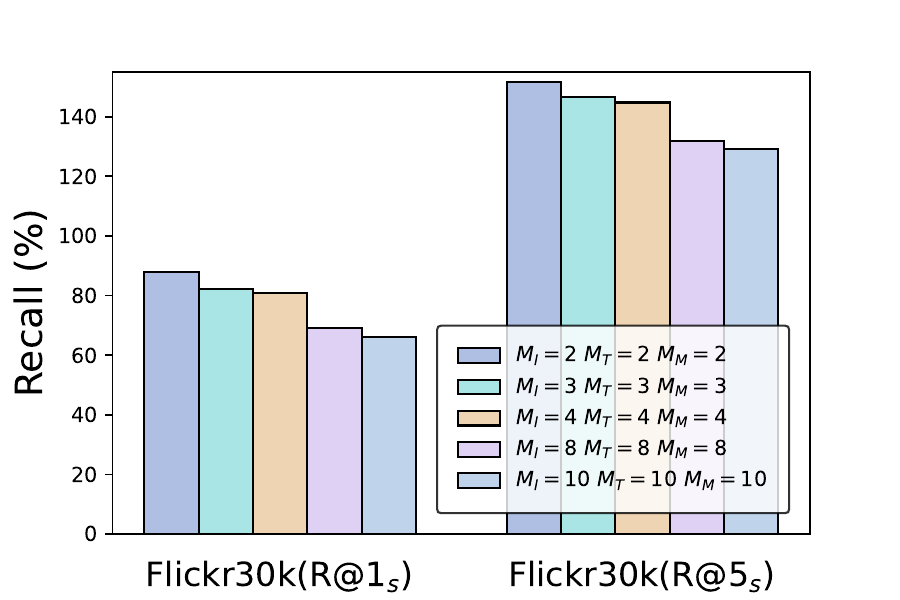}\label{fig3:subd}}
    \vspace{-0.5em}
  \caption{Average accuracy and recall of the model under different numbers of clients.}
  \label{fig-clients}
  \vspace{-2em}
\end{figure}

\noindent\textbf{The Impact of Different Numbers of Clients.}
Figure~\ref{fig-clients} illustrates the impact of varying numbers of clients on the model's average accuracy and recall.
In classification tasks, with the incremental increase in the number of clients, the classification accuracy demonstrates a certain degree of fluctuation. Nevertheless, from an overall perspective, its performance remains reasonably stable.
On the Flickr30k dataset, we observe that as the number of clients increases, retrieval performance declines. This is because more clients distribute the same dataset samples, increasing training difficulty and reducing both performance and generalization.

\noindent\textbf{Performance of AproMFL Under Different Encoders.}
Table \ref{encoder} presents the performance of AproMFL under encoders with different structures.
We employed the ViT-B/32, ViT-L/14, ViT-H/14 models as encoders. As indicated in Table \ref{encoder}, AproMFL yields superior performance with the increase in the scale of the encoder model. This phenomenon can be attributed to the fact that larger-scale models exhibit enhanced representational power and generalization capability. Specifically, when a projection head is appended to these backbone networks for fine-tuning, the model is capable of preserving the general knowledge embedded in the backbone while adapting to downstream tasks.
\begin{table}[t]
	\centering
	\begin{tabular}{c|c|ccc}
		\bottomrule
        Datasets & Metrics   & ViT-B/32 & ViT-L/14& ViT-H/14\\ \hline
	  \multirow{2}*{CIFAR-10}& Acc@1	&52.87\% & 59.71\%&62.80\%\\
        & Acc@5	&70.24\%& 67.37\%&65.79\%\\
        \hline
      \multirow{2}*{AG-NEWS}&Acc@1	&36.78\% & 59.14\%&59.15\%\\
        & Acc@2	&50.04\%& 68.32\%&69.64\%\\
      \hline
       \multirow{2}*{Flickr30k}&{R@1}$_s$& 69.59\%&89.36\% & 94.67\%\\
       &R@5$_s$& 131.35\%&150.61\% & 156.72\%\\
      
       \toprule
       
	\end{tabular}
    \vspace{-0.5em}
    \caption{The performance of AproMFL under different encoders.}
 \label{encoder}
 \vspace{-2.3em}
\end{table}

\noindent\textbf{The Impact of Modality Completion Parameter $\mathcal O$.}
The value of the modality completion parameter governs the quality of prototype semantic completion for unimodal clients. To examine its influence on model performance, we set the parameter $\mathcal O$ to 2, 5, 8, and 10, and the corresponding results are detailed in Table \ref{completion_o}.
As $\mathcal O$ increases, AproMFL performance decreases across most datasets. This is because larger $\mathcal O$ involves more multimodal prototypes for semantic completion of unimodal prototypes. Some tail multimodal prototypes, which have low similarity to unimodal ones, may introduce noise, reducing AproMFL's performance.

\noindent\textbf{The Impact of Different Types of Mapping Modules.}
Table \ref{mapping} presents the accuracy and recall of AproMFL under different mapping modules. 
For the Flickr30k, the recall of the 1-layer FCN model outperforms that of the 3-layer FCN model in both i2t and t2i retrieval tasks. For classification, the Acc@5 and Acc@2 of the 1-layer FCN model are also higher. This suggests that our method performs better with simpler mapping modules. Fewer layers lead to fewer parameters, reducing both computational time and communication overhead.
However, the Acc@1 of the 1-layer FCN is lower than that of the 3-layer FCN.
This is because we use the CLIP as the encoder and obtain the final classification model by fine-tuning the mapping and classification modules. Unlike the contrastive loss used in CLIP, our training process employs cross-entropy loss. Due to the difference in optimization objectives, more complex networks achieve better performance in fewer training epochs, while simpler networks perform worse.
\begin{table}[t]
	\centering
	\begin{tabular}{c|c|c|c}
		\bottomrule
		\multirow{2}*{Datasets} & \multirow{2}*{Metrics} &\multicolumn{2}{c}{Mapping Module}\\
        \cline{3-4}
        & &1-layer FCN & 3-layer FCN \\
        \hline
	  \multirow{2}*{CIFAR-10}& Acc@1& 56.23\%&59.71\%	\\
        & Acc@5& 75.11\%&67.37\%	\\
        \hline
      \multirow{2}*{AG-NEWS}&Acc@1& 51.14\%&59.14\%	\\
        & Acc@2& 72.24\%&68.32\%	\\
      \hline
       \multirow{4}*{Flickr30k}&{R@1}(i2t)&61.95\% & 50.09\%\\
       &R@5(i2t)&87.60\% & 78.75\%\\
       &{R@1}(t2i)&50.60\% & 39.47\%\\
       &R@5(t2i)&78.84\% & 71.86\%\\
      \hline
       
	\end{tabular}
     \vspace{-0.5em}
    \caption{Comparisons of accuracy and recall under different mapping modules.}
 \label{mapping}
 \vspace{-2em}
\end{table}

\section{Conclusions}
In this paper, we propose AproMFL, an adaptive prototype-based MFL framework for mixed modalities. 
In AproMFL, prototype construction and aggregation are performed without requiring a unified label space.
The server unifies unimodality prototypes into multimodal prototypes through semantic completion, and then derives global prototype pairs via multimodal clustering. 
To reduce performance degradation from model aggregation in heterogeneous task, we propose a relationship graph-based model aggregation method. 
Finally, clients complete local training under the guidance of global prototype knowledge transfer loss and global model knowledge transfer loss, reducing the discrepancy between local and global representations.
Experimental results show AproMFL effectively facilitates collaborative training of clients with mixed modalities and task heterogeneity.

\bibliography{aaai26}

\begin{thebibliography}{35}
\providecommand{\natexlab}[1]{#1}

\bibitem[{Bao et~al.(2024)Bao, Zhang, Miao, Gong, Hu, Liu, Liu, and Shi}]{bao2023multimodal}
Bao, G.; Zhang, Q.; Miao, D.; Gong, Z.; Hu, L.; Liu, K.; Liu, Y.; and Shi, C. 2024.
\newblock Multimodal federated learning with missing modality via prototype mask and contrast.
\newblock In \emph{nternational Conference on Machine Learning}. Vienna, Austria.

\bibitem[{Chen et~al.(2024)Chen, Zhang, Krompass, Gu, and Tresp}]{chen2024feddat}
Chen, H.; Zhang, Y.; Krompass, D.; Gu, J.; and Tresp, V. 2024.
\newblock Fed{DAT}: An approach for foundation model finetuning in multi-modal heterogeneous federated learning.
\newblock In \emph{Proceedings of the AAAI Conference on Artificial Intelligence}, 11285--11293. Vancouver, Canada.

\bibitem[{Chen and Zhang(2022)}]{chen2022fedmsplit}
Chen, J.; and Zhang, A. 2022.
\newblock Fed{MS}plit: Correlation-adaptive federated multi-task learning across multimodal split networks.
\newblock In \emph{Proceedings of the 28th ACM SIGKDD conference on knowledge discovery and data mining}, 87--96. Washington, DC, USA.

\bibitem[{Dai et~al.(2023)Dai, Chen, Li, Heinecke, Sun, and Xu}]{dai2023tackling}
Dai, Y.; Chen, Z.; Li, J.; Heinecke, S.; Sun, L.; and Xu, R. 2023.
\newblock Tackling data heterogeneity in federated learning with class prototypes.
\newblock In \emph{Proceedings of the AAAI Conference on Artificial Intelligence}, volume~37, 7314--7322.

\bibitem[{Feng et~al.(2023)Feng, Bose, Zhang, Hebbar, Ramakrishna, Gupta, Zhang, Avestimehr, and Narayanan}]{feng2023fedmultimodal}
Feng, T.; Bose, D.; Zhang, T.; Hebbar, R.; Ramakrishna, A.; Gupta, R.; Zhang, M.; Avestimehr, S.; and Narayanan, S. 2023.
\newblock Fedmultimodal: A benchmark for multimodal federated learning.
\newblock In \emph{Proceedings of the 29th ACM SIGKDD Conference on Knowledge Discovery and Data Mining}, 4035--4045. Long Beach, CA, USA.

\bibitem[{Healy and McInnes(2024)}]{Healy2024}
Healy, J.; and McInnes, L. 2024.
\newblock Uniform manifold approximation and projection.
\newblock \emph{Nature Reviews Methods Primers}, 4(1): 82.

\bibitem[{Hsu, Qi, and Brown(2019)}]{hsu2019measuring}
Hsu, T.-M.~H.; Qi, H.; and Brown, M. 2019.
\newblock Measuring the effects of non-identical data distribution for federated visual classification.
\newblock \emph{arXiv preprint arXiv:1909.06335}, pp(99).

\bibitem[{Huang et~al.(2023)Huang, Ye, Shi, Li, and Du}]{huang2023rethinking}
Huang, W.; Ye, M.; Shi, Z.; Li, H.; and Du, B. 2023.
\newblock Rethinking federated learning with domain shift: A prototype view.
\newblock In \emph{2023 IEEE/CVF Conference on Computer Vision and Pattern Recognition (CVPR)}, 16312--16322. Vancouver, BC, Canada.

\bibitem[{Krizhevsky, Hinton et~al.(2009)}]{krizhevsky2009learning}
Krizhevsky, A.; Hinton, G.; et~al. 2009.
\newblock Learning multiple layers of features from tiny images.

\bibitem[{Le et~al.(2024)Le, Thwal, Qiao, Tun, Nguyen, and Hong}]{le2024cross}
Le, H.~Q.; Thwal, C.~M.; Qiao, Y.; Tun, Y.~L.; Nguyen, M.~N.; and Hong, C.~S. 2024.
\newblock Cross-Modal Prototype based Multimodal Federated Learning under Severely Missing Modality.
\newblock \emph{arXiv preprint arXiv:2401.13898}, pp(99): 1--12.

\bibitem[{Li et~al.(2023{\natexlab{a}})Li, Li, Zhu, Cui, and Li}]{li2023prototype}
Li, J.; Li, F.; Zhu, L.; Cui, H.; and Li, J. 2023{\natexlab{a}}.
\newblock Prototype-guided knowledge transfer for federated unsupervised cross-modal hashing.
\newblock In \emph{Proceedings of the 31st ACM International Conference on Multimedia}, 1013--1022. Ottawa, ON, Canada.

\bibitem[{Li et~al.(2023{\natexlab{b}})Li, Tang, Chen, Weng, Peng, and Yang}]{li2023exploring}
Li, M.; Tang, X.; Chen, S.; Weng, Y.; Peng, L.; and Yang, W. 2023{\natexlab{b}}.
\newblock Exploring the Impact of Non-IID on Federated Learning.
\newblock In \emph{2023 International Conference on Blockchain Technology and Information Security}, 159--167. Xi'an, China.

\bibitem[{Li, He, and Song(2021)}]{li2021model}
Li, Q.; He, B.; and Song, D. 2021.
\newblock Model-contrastive federated learning.
\newblock In \emph{Proceedings of the IEEE/CVF conference on computer vision and pattern recognition}, 10713--10722. virtual.

\bibitem[{Li et~al.(2020)Li, Sahu, Zaheer, Sanjabi, Talwalkar, and Smith}]{li2020federated}
Li, T.; Sahu, A.~K.; Zaheer, M.; Sanjabi, M.; Talwalkar, A.; and Smith, V. 2020.
\newblock Federated optimization in heterogeneous networks.
\newblock In \emph{Proceedings of Machine learning and systems}, 429--450. Austin, TX, USA.

\bibitem[{Li et~al.(2024)Li, Hou, Liu, Li, Yang, Wang, Shi, Xie, Zhang, Xu et~al.}]{li2024federated}
Li, Z.; Hou, Z.; Liu, H.; Li, T.; Yang, C.; Wang, Y.; Shi, C.; Xie, L.; Zhang, W.; Xu, L.; et~al. 2024.
\newblock Federated Learning in Large Model Era: Vision-Language Model for Smart City Safety Operation Management.
\newblock In \emph{Companion Proceedings of the ACM on Web Conference 2024}, 1578--1585. Singapore, Singapore.

\bibitem[{McMahan et~al.(2017)McMahan, Moore, Ramage, Hampson, and y~Arcas}]{mcmahan2017communication}
McMahan, B.; Moore, E.; Ramage, D.; Hampson, S.; and y~Arcas, B.~A. 2017.
\newblock Communication-efficient learning of deep networks from decentralized data.
\newblock In \emph{Artificial intelligence and statistics}, 1273--1282. Fort Lauderdale, FL, {USA}.

\bibitem[{Peng, Bian, and Xu(2024)}]{peng2024fedmm}
Peng, Y.; Bian, J.; and Xu, J. 2024.
\newblock Fed{mm}: Federated Multi-Modal Learning with Modality Heterogeneity in Computational Pathology.
\newblock In \emph{ICASSP 2024-2024 IEEE International Conference on Acoustics}, 1696--1700. Seoul, Republic of Korea.

\bibitem[{Poudel et~al.(2024)Poudel, Shrestha, Amgain, Shrestha, Gyawali, and Bhattarai}]{poudel2024car}
Poudel, P.; Shrestha, P.; Amgain, S.; Shrestha, Y.~R.; Gyawali, P.; and Bhattarai, B. 2024.
\newblock {CAR-MFL}: Cross-Modal Augmentation by Retrieval for Multimodal Federated Learning with Missing Modalities.
\newblock In \emph{International Conference on Medical Image Computing and Computer-Assisted Intervention}, 102--112. Marrakesh, Morocco.

\bibitem[{Qi and Li(2024)}]{qi2024adaptive}
Qi, F.; and Li, S. 2024.
\newblock Adaptive Hyper-graph Aggregation for Modality-Agnostic Federated Learning.
\newblock In \emph{Proceedings of the IEEE/CVF Conference on Computer Vision and Pattern Recognition}, 12312--12321. Seattle, WA, USA.

\bibitem[{Radford et~al.(2021)Radford, Kim, Hallacy, Ramesh, Goh, Agarwal, Sastry, Askell, Mishkin, Clark et~al.}]{radford2021learning}
Radford, A.; Kim, J.~W.; Hallacy, C.; Ramesh, A.; Goh, G.; Agarwal, S.; Sastry, G.; Askell, A.; Mishkin, P.; Clark, J.; et~al. 2021.
\newblock Learning transferable visual models from natural language supervision.
\newblock In \emph{International conference on machine learning}, 8748--8763. Virtual Event.

\bibitem[{Sun et~al.(2024)Sun, Mendieta, Dutta, Li, and Chen}]{sun2024towards}
Sun, G.; Mendieta, M.; Dutta, A.; Li, X.; and Chen, C. 2024.
\newblock Towards Multi-modal Transformers in Federated Learning.
\newblock In \emph{European Conference on Computer Vision}, 229--246.

\bibitem[{Wang et~al.(2020{\natexlab{a}})Wang, Yurochkin, Sun, Papailiopoulos, and Khazaeni}]{wang2020federated}
Wang, H.; Yurochkin, M.; Sun, Y.; Papailiopoulos, D.; and Khazaeni, Y. 2020{\natexlab{a}}.
\newblock Federated learning with matched averaging.
\newblock In \emph{8th International Conference on Learning Representations}, 1. Addis Ababa, Ethiopia.

\bibitem[{Wang et~al.(2020{\natexlab{b}})Wang, Liu, Liang, Joshi, and Poor}]{wang2020tackling}
Wang, J.; Liu, Q.; Liang, H.; Joshi, G.; and Poor, H.~V. 2020{\natexlab{b}}.
\newblock Tackling the objective inconsistency problem in heterogeneous federated optimization.
\newblock In \emph{Advances in neural information processing systems}, 7611--7623. virtual.

\bibitem[{Wang et~al.(2025)Wang, Gai, Yu, Zhang, and Zhu}]{wang2025pravfed}
Wang, S.; Gai, K.; Yu, J.; Zhang, Z.; and Zhu, L. 2025.
\newblock PraVFed: Practical Heterogeneous Vertical Federated Learning via Representation Learning.
\newblock \emph{IEEE Transactions on Information Forensics and Security}.

\bibitem[{Wang et~al.(2024)Wang, Fu, Kanagavelu, Wei, Liu, and Goh}]{wang2024aggregation}
Wang, Y.; Fu, H.; Kanagavelu, R.; Wei, Q.; Liu, Y.; and Goh, R. S.~M. 2024.
\newblock An aggregation-free federated learning for tackling data heterogeneity.
\newblock In \emph{Proceedings of the IEEE/CVF Conference on Computer Vision and Pattern Recognition}, 26233--26242.

\bibitem[{Xiong et~al.(2023)Xiong, Yang, Song, Wang, and Xu}]{xiong2023client}
Xiong, B.; Yang, X.; Song, Y.; Wang, Y.; and Xu, C. 2023.
\newblock Client-Adaptive Cross-Model Reconstruction Network for Modality-Incomplete Multimodal Federated Learning.
\newblock In \emph{Proceedings of the 31st ACM International Conference on Multimedia}, 1241--1249.

\bibitem[{Yan et~al.(2024)Yan, Cui, Wuerkaixi, Zhang, Han, Niu, Sugiyama, and Zhang}]{yan2024balancing}
Yan, K.; Cui, S.; Wuerkaixi, A.; Zhang, J.; Han, B.; Niu, G.; Sugiyama, M.; and Zhang, C. 2024.
\newblock Balancing Similarity and Complementarity for Federated Learning.
\newblock In \emph{Proceedings of the 41st International Conference on Machine Learning}, 55739--55758. Vienna, Austria.

\bibitem[{Yazdinejad et~al.(2024)Yazdinejad, Dehghantanha, Karimipour, Srivastava, and Parizi}]{yazdinejad2024robust}
Yazdinejad, A.; Dehghantanha, A.; Karimipour, H.; Srivastava, G.; and Parizi, R.~M. 2024.
\newblock A robust privacy-preserving federated learning model against model poisoning attacks.
\newblock \emph{IEEE Transactions on Information Forensics and Security}, 19: 6693--6708.

\bibitem[{Young et~al.(2014)Young, Lai, Hodosh, and Hockenmaier}]{young2014image}
Young, P.; Lai, A.; Hodosh, M.; and Hockenmaier, J. 2014.
\newblock From image descriptions to visual denotations: New similarity metrics for semantic inference over event descriptions.
\newblock \emph{Transactions of the Association for Computational Linguistics}, 2: 67--78.

\bibitem[{Yu et~al.(2023)Yu, Liu, Wang, Xu, and Liu}]{yu2023multimodal}
Yu, Q.; Liu, Y.; Wang, Y.; Xu, K.; and Liu, J. 2023.
\newblock Multimodal federated learning via contrastive representation ensemble.
\newblock In \emph{The Eleventh International Conference on Learning Representations}. Kigali, Rwanda.

\bibitem[{Zhang et~al.(2024)Zhang, Liu, Hua, and Cao}]{zhang2024fedtgp}
Zhang, J.; Liu, Y.; Hua, Y.; and Cao, J. 2024.
\newblock Fed{TGP}: Trainable global prototypes with adaptive-margin-enhanced contrastive learning for data and model heterogeneity in federated learning.
\newblock In \emph{Proceedings of the AAAI conference on artificial intelligence}, volume~38, 16768--16776.

\bibitem[{Zhang, Zhao, and LeCun(2015)}]{zhang2015character}
Zhang, X.; Zhao, J.; and LeCun, Y. 2015.
\newblock Character-level convolutional networks for text classification.
\newblock In \emph{Advances in Neural Information Processing Systems 28: Annual Conference on Neural Information Processing Systems}, 649--657. Montreal, Quebec, Canada.

\bibitem[{Zhao, Barnaghi, and Haddadi(2022)}]{zhao2022multimodal}
Zhao, Y.; Barnaghi, P.; and Haddadi, H. 2022.
\newblock Multimodal federated learning on iot data.
\newblock In \emph{2022 IEEE/ACM Seventh International Conference on Internet-of-Things Design and Implementation}, 43--54.

\bibitem[{Zhou et~al.(2025)Zhou, Qu, You, Zhou, Tang, Zheng, Cai, and Wu}]{zhou2025fedsa}
Zhou, Y.; Qu, X.; You, C.; Zhou, J.; Tang, J.; Zheng, X.; Cai, C.; and Wu, Y. 2025.
\newblock Fedsa: A unified representation learning via semantic anchors for prototype-based federated learning.
\newblock In \emph{Proceedings of the AAAI Conference on Artificial Intelligence}, volume~39, 23009--23017.

\bibitem[{Zong et~al.(2021)Zong, Xie, Zhou, Wu, Zhang, and Xu}]{zong2021fedcmr}
Zong, L.; Xie, Q.; Zhou, J.; Wu, P.; Zhang, X.; and Xu, B. 2021.
\newblock {FedCMR}: Federated cross-modal retrieval.
\newblock In \emph{Proceedings of the 44th International ACM SIGIR Conference on Research and Development in Information Retrieval}, 1672--1676. Virtual Event, Canada.

\end{thebibliography}
\end{document}